\documentclass{article}

\PassOptionsToPackage{numbers}{natbib} %
\usepackage[dblblindworkshop, final]{neurips_2026}
\workshoptitle{Foundation Models for Temporal Systems: From Forecasting to World Modeling}

\usepackage[utf8]{inputenc} %
\usepackage[T1]{fontenc}    %
\usepackage{courier}        %
\usepackage{hyperref}       %
\usepackage{url}            %
\usepackage{booktabs}       %
\usepackage{amsfonts}       %
\usepackage{amsmath}
\usepackage{amsthm}
\usepackage{nicefrac}       %
\usepackage{microtype}      %
\usepackage{xcolor}         %
\usepackage{graphicx}
\usepackage{enumitem}
\usepackage{multirow}
\usepackage{caption}
\usepackage{wrapfig}


\newcommand{\algname}{QUAL}%
\newtheorem*{problem}{Problem Definition}

\title{\algname{}: Quantile Alignment Improves Zero-Shot Time-Series Foundation Models}

\author{%
  \textbf{Junhyeok Kang, Sangjun Han, Jun Seo, Seunghan Lee, Jaehoon Lee, Dongwan Kang,} \\[0.2em]
  \textbf{Minjae Kim, Sungdong Yoo, Tae Yoon Lim, Hwanil Choi, Soonyoung Lee, Wonbin Ahn} \\[0.6em]
  LG AI Research
}

\begin{document}

\maketitle

\begin{abstract}
Time-series foundation models\,(TSFMs) provide zero-shot forecasts for unseen series, but their accuracy varies across datasets. Combining several models can reduce this variability, yet zero-shot deployment provides no target labels for selecting or weighting them. We find that a model pool is useful not only as a final ensemble, but also as a probability reference for transforming each member forecast. Aligning quantiles through this pool-relative reference improves every member forecast in aggregate. Motivated by this finding, we propose \algname{}\,(\textbf{Qu}antile \textbf{al}ignment), a simple label-free framework for improving and combining diverse TSFMs. \algname{} first produces aligned member forecasts, then applies Huber aggregation and rearrangement to obtain a robust combined forecast without training or model modification. On GIFT-Eval, quantile alignment reduces average member MASE and CRPS by $7.4\%$ and $9.8\%$. The full framework remains robust when weak forecasters enter a candidate pool whose quality cannot be verified in advance. Code is available at \url{https://github.com/junhyeokkang/QUAL}.
\end{abstract}

\section{Introduction}

Time-series forecasting is a critical task across diverse real-world domains, from supply chain management to finance\,\citep{trirat2026universal, kang2025vardrop, kang2026channel, kang2026traceable}. Time-series foundation models\,(TSFMs) can forecast series from unseen datasets without additional training \citep{das2024timesfm,auer2025tirex, ansari2025chronos2, lee2026exaone}. Their accuracy varies across datasets and forecast horizons, so no single model is reliably best \citep{aksu2024gifteval}. For a new series, observed target values are unavailable to identify the strongest model or learn combination weights. A model pool can reduce this selection risk, but its forecasts must be combined without target feedback. For the same reason, the reliability of the candidate pool itself cannot be verified in advance.

Most TSFMs express uncertainty through a finite set of quantiles.
Label-free rules usually combine these forecasts directly. Quantile averaging
uses only values reported at the same nominal probability levels, discarding how
those values are positioned relative to the other member distributions. The linear opinion pool uses reconstructed distributions but immediately collapses the pool into a single forecast. Consequently, this approach is highly vulnerable when weak forecasters are included, which poses a key risk in label-free settings where model quality cannot be pre-verified.

\begin{figure}[t]
\centering
\includegraphics[width=\linewidth]{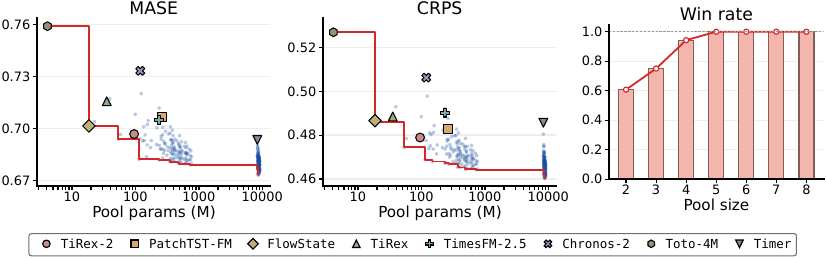}
\caption{Parameter efficiency and subset win rate on GIFT-Eval. MASE and
CRPS panels plot each pool against total parameter count; markers are
single TSFMs, red lines are Pareto frontiers. Win rate is the fraction of
subsets that beat the best individual model on both metrics.}
\label{fig:teaser}
\end{figure}

In this paper, we propose \algname{}\,(\textbf{Qu}antile \textbf{al}ignment), a label-free framework that uses the
pooled distribution as a probability reference instead of a forecast. The
pooled cumulative distribution function reflects the average pool-relative
rank of any value, so it can serve as a shared probability reference
without any labels --- the same object that the linear opinion pool returns
as a forecast. \algname{} keeps this reference but does not collapse the
pool: it re-expresses each member on the shared probability reference and
passes the resulting aligned forecasts to a robust aggregator downstream.
Because the members remain separate, their disagreement stays visible when
the pool contains a weak forecaster and the aggregator can limit its
influence rather than average it in.

Figure~\ref{fig:teaser} shows the aggregate effect. Across multi-model
subsets of the strong pool, \algname{} beats the best individual TSFM
on both MASE and CRPS in $91.8\%$ of cases when the subset has at least
three members, and always when it has at least five. %

Our contributions are:
\begin{itemize}[leftmargin=1.2em, itemsep=1pt, topsep=2pt]
    \item We identify the pooled cumulative distribution as a label-free probability reference and show that using it to re-express each member forecast, rather than as the forecast itself, improves every evaluated TSFM in aggregate, including under leave-one-out references.
    \item We propose \algname{}, which pairs this alignment with Huber
    aggregation and rearrangement to obtain a robust combined forecast
    without training or model updates.
    \item We show empirically that keeping the aligned members separate is
    what makes the combined forecast robust when weak forecasters are
    silently added to the pool.
\end{itemize}

\section{Related Work}

Zero-shot TSFMs span diverse architectures, including decoder-only transformers \citep{das2024timesfm,rasul2023lagllama}, tokenized language modeling \citep{ansari2024chronos}, group attention across related series \citep{ansari2025chronos2}, cross-frequency universal designs \citep{woo2024moirai}, and long-context in-context learning \citep{auer2025tirex}, motivating model coordination rather than single-model selection. Recent work on TSFM combination takes several routes: Toto-2.0-FnF trains a learned gate \citep{toto2026fnf}, CastStar and TimeCopilot wrap models in an agent that selects or ensembles them per query \citep{caststar2026,garza2025timecopilot}, conversational forecasters delegate the choice to an LLM judge \citep{cao2025conversational}, and Cobra Agent runs an internal validation pipeline before returning a forecast \citep{cobra2026agent}. However, each of these approaches needs signal outside the per-request quantile forecasts --- a labelled training corpus, external agent tooling, or an internal validation split. Classical forecast combination rules average either values at fixed probability levels or reconstructed distributions \citep{timmermann2006forecast,lichtendahl2013better,busetti2017quantile}, whereas weighted variants depend on past performance or explicit error assumptions \citep{bates1969combination,dawid1979maximum,guha2024smoothie}. In contrast, \algname{} operates under a strictly label-free constraint, requiring only current-request quantile forecasts without relying on target labels or historical supervision.

\section{Proposed Method: \algname{}}
\label{sec:method}

\begin{problem}[Label-Free Forecast Combination]
Given an input lookback window $\mathbf x\in\mathbb{R}^{L\times C}$ and $K$
pretrained forecasters, let
$\boldsymbol{\tau}=(\tau_1,\ldots,\tau_Q)\in(0,1)^Q$ be ordered probability
levels with $0<\tau_1<\cdots<\tau_Q<1$. For a target horizon and variable
conditioned on $\mathbf x$, model $k$ returns a quantile vector
$\mathbf q_k=(q_k(\tau_1),\ldots,q_k(\tau_Q))\in\mathbb{R}^{Q}$. The task is to
construct aligned member forecasts
$\widetilde{\mathbf q}_k\in\mathbb{R}^{Q}$ and a combined forecast
$\hat{\mathbf q}\in\mathbb{R}^{Q}$ satisfying
$\hat q(\tau_1)\le\cdots\le\hat q(\tau_Q)$, without access to target-domain
labels or updates to the pretrained models. The procedure is applied
independently to each horizon and variable.
\end{problem}

\subsection{Overall Framework}
To obtain reliable forecasts from diverse TSFMs without target labels,
\algname{} separates transformation from aggregation. It first reconstructs
each member CDF, forms a pooled CDF, and re-expresses each member's reported
quantiles through this pool-relative reference. It then robustly aggregates
the aligned member forecasts and rearranges the result to ensure noncrossing
quantiles.

\subsection{Quantile Alignment}
\label{sec:align}

\noindent\textbf{Pooled CDF Construction.}
Level-wise averaging ignores how each reported value ranks against the
other member distributions. \algname{} uses this cross-model rank
information: for each member, we sort $\mathbf q_k\in\mathbb{R}^{Q}$ and
linearly interpolate $(q_k(\tau_j),\tau_j)$ into a piecewise-linear CDF
$G_k$. Their equally weighted mixture $\bar G(x)=\frac{1}{K}\sum_{k=1}^{K}
G_k(x)$ is the pooled distribution function, so $\bar G(x)$ is the average
percentile assigned to $x$ by the pool and serves as the label-free
probability reference. Because finite quantiles do not identify the tails,
we clip $G_k$ outside its reported range.

\noindent\textbf{Member-wise Quantile Alignment.}
Alignment transforms each member forecast before final aggregation. For model $k$, the pooled percentiles $\mathbf a_k=(a_k(\tau_1),\ldots,a_k(\tau_Q))\in[0,1]^Q$ are $a_k(\tau_j)=\bar G(q_k(\tau_j))$, which relabels each value by its rank in the pool. Linear interpolation of the rank--value pairs $(a_k(\tau_j),q_k(\tau_j))$ defines $\widetilde Q_k:[0,1]\to\mathbb{R}$. Evaluating this function at the requested probability levels gives $\widetilde{\mathbf q}_k=(\widetilde Q_k(\tau_1),\ldots, \widetilde Q_k(\tau_Q))\in\mathbb{R}^{Q}$. Each member is thereby returned on a common probability reference while retaining member-specific forecast structure. Levels outside the aligned range are clipped to the nearest endpoint.

\subsection{Robust Aggregation and Rearrangement}
\label{sec:aggregate}

\noindent\textbf{Huber Robust Aggregation.}
After alignment, the combined forecast should not be dominated by atypical members. Since target labels are unavailable, weak members cannot be removed by validation and a simple mean can be fragile. We therefore stack the aligned values into $\mathbf Z=[z_{kj}]\in\mathbb{R}^{K\times Q}$, where $z_{kj}=\widetilde q_k(\tau_j)$, and combine them with a Huber robust center, which behaves like an average near the consensus but limits large deviations. At level $\tau_j$, let $\bar z_j=\operatorname{median}_k z_{kj}$ and define the robust scale $s_j=b_{\mathrm{MAD}}\operatorname{median}_k|z_{kj}-\bar z_j|+\epsilon$, where the median term is the median absolute deviation\,(MAD), $b_{\mathrm{MAD}}>0$ is a scale-calibration constant, and $\epsilon$ handles exact agreement. The robust center is $h_j=\arg\min_{m}\sum_{k=1}^{K}\rho_c((z_{kj}-m)/s_j)$, where $\rho_c$ is the Huber loss with threshold $c>0$ \citep{huber1964robust}. Collecting the level-wise centers gives $\mathbf h=(h_1,\ldots,h_Q)\in\mathbb{R}^{Q}$.

\noindent\textbf{Disagreement-based Influence.}
\algname{} applies robustness locally rather than assigning one reliability score to each model. The disagreement of member $k$ at level $\tau_j$ is its standardized distance from $h_j$, $d_{kj}=|z_{kj}-h_j|/s_j$. Collecting these values gives a disagreement matrix $\mathbf D=[d_{kj}]\in\mathbb{R}_{\ge0}^{K\times Q}$. The effective Huber weight matrix $\mathbf W=[w_{kj}]\in[0,1]^{K\times Q}$ has $w_{kj}=1$ for $d_{kj}\le c$ and $w_{kj}=c/d_{kj}$ otherwise. Thus, a member keeps full influence where it agrees with the robust center and is downweighted only where its disagreement is large.

\noindent\textbf{Quantile Rearrangement.}
Finally, the combined forecast must remain a valid quantile function. Since the $Q$ robust centers are estimated independently, they can occasionally cross. \algname{} restores a valid order by setting $\hat{\mathbf q}\in\mathbb{R}^{Q}$ to the sorted values of $\mathbf h=(h_1,\ldots,h_Q)$. %

\section{Evaluation}

\subsection{Experimental Setup}
\label{sec:setup}

\noindent\textbf{Pool Construction.}
Our forecaster pool separates \emph{strong} and \emph{weak} members. The
strong pool consists of eight public TSFMs spanning architectures and
scales: TiRex \citep{auer2025tirex} and TiRex-2 \citep{podest2026tirex2generalizingtirexmultivariate},
PatchTST-FM \citep{wen2026generic}, FlowState \citep{graf2025flowstate},
TimesFM-2.5 \citep{das2024timesfm}, Timer-S1 \citep{liu2026timer},
Chronos-2 \citep{ansari2025chronos2}, and
Toto-2.0-4M \citep{khwaja2026toto2}. We also consider four weak
forecasters used to test robustness in Section~\ref{sec:aggregation}: the
GIFT-Eval benchmark's seasonal-naive reference, last value, a Gaussian
fit, and the history mean.

\noindent\textbf{Dataset and Metrics.}
We evaluate all methods on the GIFT-Eval
benchmark \citep{aksu2024gifteval} with mean absolute scaled error\,(MASE) and continuous ranked probability score\,(CRPS). We report geometric-mean scores relative to a seasonal-naive forecast, following the benchmark protocol \citep{aksu2024gifteval}.

\begin{table}[t]
    \setlength{\tabcolsep}{3.7pt}
    \centering
    \small
    \caption{Per-member GIFT-Eval accuracy before (\emph{Origin.}) and after
    (\emph{Align.}) quantile alignment; \emph{LOO}\,(leave-one-out) removes the evaluated model
    from the pooled reference. $\Delta^{\text{Align}}_{\text{Imp}}$ and
    $\Delta^{\text{LOO}}_{\text{Imp}}$ are the percent improvements of
    \emph{Align.}\ and \emph{LOO}\ over \emph{Origin.}
    Every member improves on both metrics.}
    \label{tab:alignment}
    \begin{tabular}{lccrcrccrcr}
    \toprule
     & \multicolumn{5}{c}{MASE\,($\downarrow$)} & \multicolumn{5}{c}{CRPS\,($\downarrow$)} \\
    \cmidrule(lr){2-6}\cmidrule(lr){7-11}
    Model & Origin. & Align. & $\Delta^{\text{Align}}_{\text{Imp}}$ & LOO & $\Delta^{\text{LOO}}_{\text{Imp}}$ & Origin. & Align. & $\Delta^{\text{Align}}_{\text{Imp}}$ & LOO & $\Delta^{\text{LOO}}_{\text{Imp}}$ \\
    \midrule
    TiRex-2\,\citep{podest2026tirex2generalizingtirexmultivariate}        & 0.6969 & 0.6810 & $2.3\%$  & 0.6818 & $2.2\%$  & 0.4789 & 0.4704 & $1.8\%$  & 0.4716 & $1.5\%$  \\
    PatchTST-FM\,\citep{wen2026generic}    & 0.7069 & 0.6835 & $3.3\%$  & 0.6850 & $3.1\%$  & 0.4829 & 0.4732 & $2.0\%$  & 0.4741 & $1.8\%$  \\
    Timer-S1\,\citep{liu2026timer}       & 0.6935 & 0.6783 & $2.2\%$  & 0.6811 & $1.8\%$  & 0.4856 & 0.4748 & $2.2\%$  & 0.4764 & $1.9\%$  \\
    FlowState\,\citep{graf2025flowstate}      & 0.7015 & 0.6824 & $2.7\%$  & 0.6836 & $2.6\%$  & 0.4866 & 0.4738 & $2.6\%$  & 0.4747 & $2.4\%$  \\
    TiRex\,\citep{auer2025tirex}          & 0.7158 & 0.6838 & $4.5\%$  & 0.6840 & $4.4\%$  & 0.4885 & 0.4743 & $2.9\%$  & 0.4748 & $2.8\%$  \\
    TimesFM-2.5\,\citep{das2024timesfm}    & 0.7050 & 0.6828 & $3.2\%$  & 0.6837 & $3.0\%$  & 0.4903 & 0.4753 & $3.1\%$  & 0.4759 & $2.9\%$  \\
    Chronos-2\,\citep{ansari2025chronos2}      & 0.7333 & 0.6833 & $6.8\%$  & 0.6838 & $6.7\%$  & 0.5062 & 0.4774 & $5.7\%$  & 0.4775 & $5.7\%$  \\
    Toto-2.0-4M\,\citep{khwaja2026toto2}    & 0.7591 & 0.6859 & $9.6\%$  & 0.6834 & $10.0\%$ & 0.5269 & 0.4849 & $8.0\%$  & 0.4831 & $8.3\%$  \\
    Seasonal naive & 1.0000 & 0.7528 & $24.7\%$ & 0.7522 & $24.8\%$ & 1.0000 & 0.6569 & $34.3\%$ & 0.5769 & $42.3\%$ \\
    \midrule
    \textbf{\textit{Average}}            & \textbf{0.7458}          & \textbf{0.6904} & $\textbf{7.4\%}$ & \textbf{0.6910} & $\textbf{7.4\%}$ & \textbf{0.5495}          & \textbf{0.4957} & $\textbf{9.8\%}$ & \textbf{0.4872} & $\textbf{11.3\%}$ \\
    \midrule
    \textbf{\algname{}}\,(Ours) & -    & \textbf{0.6793} & -     & -     & -    & -     & \textbf{0.4646} & -    & -    & -      \\
    \bottomrule
    \end{tabular}
\end{table}

\subsection{Forecast Enhancement via Quantile Alignment}
\label{sec:align-effect}
Table~\ref{tab:alignment} covers the eight strong TSFMs and the seasonal-naive reference as a weak baseline to test whether our quantile alignment improves each member forecast before any target-specific selection. Alignment reduces average member MASE and CRPS by $7.4\%$ and $9.8\%$ and improves \emph{every} member on \emph{both} metrics, from the strong TSFMs down to the seasonal-naive reference (whose scores drop by $24.7\%$ and $34.3\%$). Leave-one-out (LOO) scores, obtained after removing the evaluated model from the pooled reference, are nearly identical, so the gain is not a self-inclusion artifact. The full framework further beats every zero-shot system in the pinned GIFT-Eval snapshot.

\subsection{Robustness to Weak Forecasters}
\label{sec:aggregation}
\setlength{\intextsep}{0pt}
\begin{wrapfigure}[12]{r}{0.5\linewidth}
\centering
\includegraphics[width=\linewidth]{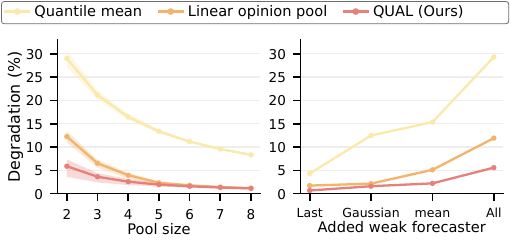}
\caption{CRPS degradation when weak forecasters enter the pool.}
\label{fig:robust}
\end{wrapfigure}

On the pool with \emph{strong} TSFMs, the linear opinion pool is still accurate, but this favorable condition is unavailable in the zero-shot setting, where the pool cannot be verified against labels. When three weak forecasters --- last value, a Gaussian fit, and the history mean --- are added to the strong pool (Figure~\ref{fig:robust}, right), \algname{} degrades by only $5.6\%$, versus $11.9\%$ for the linear opinion pool, and degrades less on $93$ of the $97$ GIFT-Eval configurations. Across subsets of the strong pool with seasonal naive added (Figure~\ref{fig:robust}, left), \algname{} has the lower CRPS on all $154$ subsets of three to five members and on $39$ of the $56$ six-member subsets. When \algname{} wins, it wins by up to $0.037$ CRPS; when it loses, by no more than $0.001$.

\section{Discussion and Conclusion}

\algname{} uses the model pool as a probability reference before ensembling. Re-expressing member forecasts through this reference consistently improves individual models, while preserving separate aligned forecasts allows Huber aggregation to protect the combined forecast against weak models. Requiring no retraining, historical records, or signals beyond current quantile forecasts, \algname{} serves as a practical default for combining heterogeneous TSFMs without label feedback.

\bibliographystyle{plainnat}
\bibliography{references}

\appendix
\newpage
\section{Supplementary Results}
\label{sec:appendix}

\setlength{\intextsep}{12pt}

The supplementary experiments provide the comparisons and diagnostics behind
the claims in the main paper. They use the same GIFT-Eval configurations, model
pool, and label-free inference protocol unless stated otherwise.

\subsection{Implementation Details}
\label{sec:impl}

All methods use the nine quantile levels specified by GIFT-Eval,
$\{0.1,0.2,\ldots,0.9\}$. We set the Huber threshold to $c=1.345$ and
perform two reweighting steps. For the robust scale we use
$b_{\mathrm{MAD}}=1/\Phi^{-1}(0.75)\approx1.4826$, the standard factor
that makes the MAD consistent for Gaussian standard
deviation \citep{rousseeuw1993alternatives}.

\subsection{Component Ablation}
\label{sec:ablation-table}

Table~\ref{tab:main} isolates the two components of \algname{} on the
strong pool. Alignment carries nearly all of the clean-pool accuracy
gain: applying it before mean aggregation ($0.4646$ CRPS) already
outperforms Huber on raw quantiles ($0.4657$). Huber shifts clean-pool
accuracy only marginally on top of alignment, but is retained for the
robustness role tested in Section~\ref{sec:aggregation}.

\begin{table}[h]
\centering
\small
\setlength{\tabcolsep}{5pt}
\caption{Component ablation of \algname{} on the strong pool. Each
row toggles quantile alignment and Huber robust aggregation on or off
while sharing the same TSFM forecasts; scores are averaged over the $97$
GIFT-Eval configurations.}
\label{tab:main}
\begin{tabular}{lcccc}
\toprule
Method & Quantile alignment & Huber robust aggregation & MASE\,($\downarrow$) & CRPS\,($\downarrow$) \\
\midrule
Quantile mean            & $\times$   & $\times$   & 0.6860 & 0.4677 \\
Huber aggregation        & $\times$   & \checkmark & 0.6830 & 0.4657 \\
Alignment $+$ mean       & \checkmark & $\times$   & 0.6795 & 0.4646 \\
\midrule
\algname{}\,(Ours)       & \checkmark & \checkmark & \textbf{0.6793} & \textbf{0.4646} \\
\bottomrule
\end{tabular}
\end{table}

\subsection{Aligned Forecast Examples}
\label{sec:member-gain}

\begin{figure}[h]
\centering
\includegraphics[width=0.72\linewidth]{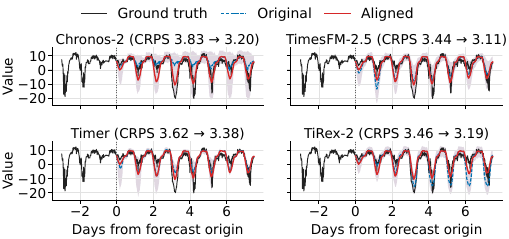}
\caption{Alignment examples on a selected ETT1 window. Shading denotes the
$10$--$90\%$ prediction intervals; solid lines are the reported medians.}
\label{fig:alignment-examples}
\end{figure}

Table~\ref{tab:alignment} in Section~\ref{sec:align-effect} reports the
member-wise alignment gain. Figure~\ref{fig:alignment-examples} shows how
the transformation looks on a selected ETT1 window. The aligned intervals
track the ground truth more tightly than the model's original output while
preserving its member-specific shape.

\newpage
\subsection{Label-Based Weighting}
\label{sec:factorial}

Learned weights and the aggregation rule can both affect forecast accuracy.
Comparing a weighted quantile average with an equally weighted linear opinion
pool changes both factors at once. Table~\ref{tab:factorial} crosses the two
choices on a six-TSFM pool of TiRex-2, PatchTST-FM, FlowState, TiRex,
TimesFM-2.5, and Chronos-2. The fitted weights are normalized inverse backtest losses estimated
separately for each configuration. This fit pools observed targets across all
series of a dataset, which the deployment setting does not provide.

Fitted weights improve both aggregation families. They reduce CRPS from
$0.4672$ to $0.4642$ for the quantile average and from $0.4634$ to $0.4604$ for
the linear opinion pool. The comparison isolates the value of labeled
model-performance records. It does not contradict the zero-shot setting, where
those records are unavailable and equal weights must be fixed before the
request.

\begin{table}[h]
\centering
\caption{CRPS for equal or fitted weights crossed with quantile averaging or the
linear opinion pool on the $97$ GIFT-Eval configurations, using the six-TSFM pool.}
\label{tab:factorial}
\small
\begin{tabular}{lrr}
\toprule
 & equal weights & fitted weights \\
\midrule
quantile average & 0.4672 & 0.4642 \\
linear opinion pool & 0.4634 & 0.4604 \\
\bottomrule
\end{tabular}
\end{table}

\subsection{Temporal Transfer of Model Skill}
\label{sec:transfer}

\begin{figure}[h]
\centering
\includegraphics[width=0.72\linewidth]{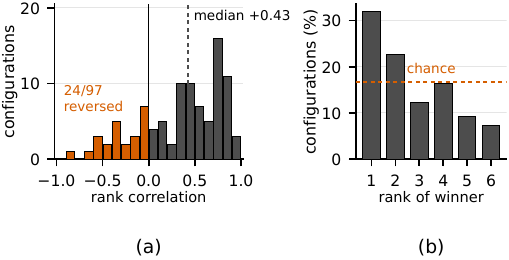}
\caption{Transfer of measured skill across the $97$ GIFT-Eval configurations,
computed on a six-TSFM diagnostic subset. (a)~Rank correlation between
backtest and forecast losses; (b)~forecast-period rank of the backtest
winner. The dashed baseline marks random selection.}
\label{fig:transfer}
\end{figure}

Fitted weights (Section~\ref{sec:factorial}) are useful only when past model
skill transfers to the unseen forecast period. We test this assumption
directly on a six-TSFM diagnostic subset of the pool: for every configuration
we rank the members by backtest and forecast-period losses and correlate the
rankings (Figure~\ref{fig:transfer}).

The rankings agree only weakly. The median Spearman correlation is $+0.43$
with quartiles from $+0.03$ to $+0.71$, and the correlation is negative in
$24$ of the $97$ configurations. The backtest winner is only the
forecast-period winner in $31$ of the $97$ configurations, against $17\%$
expected by chance. Where the rankings agree strongly (correlation
exceeding $0.5$, $42$ of $97$ configurations), fitted weights improve $40$
of $42$ cases; where the rankings are reversed, fitting weights is a net
loss ($1.0035$ CRPS ratio, only $3$ of $24$ cases improve). A practitioner
cannot distinguish these regimes without observing the targets.

\newpage
\subsection{Pool Composition}
\label{sec:composition}

\begin{figure}[h]
\centering
\includegraphics[width=0.72\linewidth]{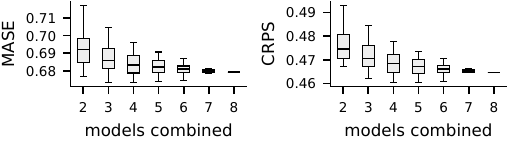}
\caption{Zero-shot accuracy by pool size across all $247$ subsets of the
strong pool, combined by \algname{}. Boxes span subset quartiles;
lines mark the median over the $97$ GIFT-Eval configurations.}
\label{fig:landscape}
\end{figure}

A practitioner may have access to only a subset of the evaluated TSFMs and
cannot validate that subset without target labels. We therefore score all
$247$ multi-model subsets of the strong pool. Under \algname{}, $218$
subsets beat the most accurate individual model in both aggregate MASE and
CRPS. The win rate rises monotonically with pool size:
$61\%$ at two members, $75\%$ at three, $94\%$ at four, and $100\%$ from
five members onward (Figure~\ref{fig:landscape}). Quantile alignment
improves every individual member, and this comparison indicates that it is
a key reason \algname{} depends less on choosing a particular pool.

\subsection{Weak-Member Contamination}
\label{sec:weak-members}

Section~\ref{sec:aggregation} tests three weak forecasters that can
plausibly enter a label-free pool despite being inaccurate. The last-value
member repeats the final context value over the forecast horizon. The
history-mean member repeats the context mean. The Gaussian member reports
quantiles of a normal distribution whose mean and standard deviation are
estimated from the context. None of these constructions uses target values
from the forecast period.

Scored alone against seasonal naive, last value reaches $1.0726$ CRPS, the
history mean $1.5855$ and the Gaussian $1.2286$, all far worse than any
TSFM in the pool (Table~\ref{tab:alignment} in
Section~\ref{sec:align-effect}). The history-mean and Gaussian forecasters
share a MASE of $1.7959$ because both place their median at the history
mean.

Adding each weak member individually to the strong pool, and then all three
together, hurts every combination rule. With all three added, CRPS rises by
$11.9\%$ under the linear opinion pool and by $5.6\%$ under \algname{}
(reported in Section~\ref{sec:aggregation}). This test measures how each
rule responds when the pool contains forecasts that are systematically less
informative than its original members.

\subsection{Limitations}
\label{sec:limitations}

\algname{} requires at least two pool members with reported quantiles at
the same nominal probability levels; a singleton pool provides no
cross-model rank information and the method reduces to the model's own
forecast. Because the pooled CDF is built from finitely many reported
quantiles, alignment is defined only inside each member's reported range
and \algname{} does not extrapolate beyond the outermost quantiles. Weak
pool members still degrade combined accuracy, but by less than under
alternative rules (Section~\ref{sec:aggregation}); the framework does not
remove the need for a reasonable candidate pool. Our evaluation covers the
$97$ GIFT-Eval configurations, and behavior on domains with substantially
different quantile calibration or reporting conventions is not measured.

\subsection{Compute Resources}
\label{sec:compute}

Every experiment is inference-only. Each of the eight strong TSFMs
generates one set of quantile forecasts per GIFT-Eval configuration on a
single NVIDIA L40S GPU\,($46$\,GB) with the checkpoints and environments
provided by their authors; producing the full pool over the $97$
configurations takes on the order of a hundred GPU-hours in aggregate,
dominated by the largest checkpoints. \algname{}'s alignment and Huber
aggregation run on CPU and finish in seconds per configuration. No model
is trained or fine-tuned, so no training compute is reported.

\subsection{Licenses of Existing Assets}
\label{sec:licenses}

We use the GIFT-Eval benchmark \citep{aksu2024gifteval} together with the
public checkpoints of eight time-series foundation models
\citep{auer2025tirex,podest2026tirex2generalizingtirexmultivariate,wen2026generic,graf2025flowstate,das2024timesfm,
liu2026timer,ansari2025chronos2,khwaja2026toto2}. Each asset is used under
the open-source license posted by its authors on the corresponding release
page; the majority of the checkpoints are distributed under Apache-2.0.
Our use is limited to running zero-shot inference for research evaluation,
which is consistent with the terms of these licenses. The four
simple weak forecasters (seasonal naive, last value, Gaussian fit, history
mean) are implemented from scratch in a few lines and do not incorporate
third-party assets.

\subsection{Broader Impacts}
\label{sec:broader-impacts}

Zero-shot time-series forecasting supports downstream decisions in
operations, energy, transportation, and financial planning. A label-free
framework that combines several TSFMs may reduce brittle single-model
choices in these settings, particularly for organizations without labelled
validation data on the target series. \algname{} does not introduce new
generative capabilities beyond those of its member models, requires no new
data collection, and does not target any group of users. As with any
probabilistic forecasting system, downstream users should not treat a
prediction interval as a substitute for domain judgement in
high-consequence decisions.

\end{document}